\pdfoutput=1
\documentclass[journal]{IEEEtran}
\usepackage[T1]{fontenc}
\usepackage[utf8]{inputenc}
\usepackage{amsmath,amsfonts,amssymb}
\usepackage{array}
\usepackage{textcomp}
\usepackage{stfloats}
\usepackage{url}
\usepackage{graphicx}
\usepackage{cite}
\usepackage{booktabs}
\usepackage[hidelinks]{hyperref}
\usepackage{orcidlink}
\newcolumntype{L}[1]{>{\raggedright\arraybackslash}p{#1}}
\newcolumntype{C}[1]{>{\centering\arraybackslash}p{#1}}
\begin{document}

\title{SHAKTI: Evidence-Grounded Conversational AI for Nutrition and Health}

\author{Chaitanya~Pallerla\,\orcidlink{0000-0003-1211-9036}$^{1,2}$
and~Dongyi~Wang\,\orcidlink{0000-0002-1224-5529}$^{1,2,*}$%

\thanks{$^{1}$Department of Biological and Agricultural Engineering, University of Arkansas, Fayetteville, AR 72701, USA.}%
\thanks{$^{2}$Department of Food Science, University of Arkansas, Fayetteville, AR 72701, USA.}%
\thanks{*Corresponding author: Dongyi Wang (e-mail: dongyiw@uark.edu).}}

\markboth{}{}

\maketitle

\begin{abstract}
Large language models (LLMs) answer many diet and health questions but can state unsupported claims and cite sources that do not support them. We present SHAKTI (Smart Health Advisor for Knowledge-based Tailored Interventions), a conversational assistant that answers only from retrieved scholarly papers and exact food-composition data, and evaluate it with retrieval, generation, grounding, efficiency and context-awareness metrics.  We indexed 484,368,086 OpenAlex works, searched exactly on a GPU for evaluation (10,065 NFCorpus, BioASQ, SciFact and PubMedQA questions) and through a compressed 230-million-abstract index in deployment. On 1,000 stratified questions, we compared single-shot, multi-hop, corrective and deep-research retrieval-augmented generation (RAG) with Claude Sonnet. SHAKTI adds conversational query rewriting, USDA FoodData Central lookups, source-link checks and online verification of quotes and numbers.  It was tested in 36 scripted conversational turns, a 248-question memory-depth test and a 38-question load test. Claims were checked by a Claude Opus judge against exactly the text the system saw.  MedCPT reranking of hybrid retrieval gave the best MRR (0.21) and MAP@100 (0.14). Across existing RAG methods, faithfulness was 0.95-0.96, claim hallucination 0.5-0.8\% and context precision up to 0.98; the deep-research agent answered 73.2\% of BioASQ questions correctly. In conversation, SHAKTI reached a context-awareness score of 97.1\%, faithfulness 95.5\%, 1.9\% unsupported claims, no contradicted claims and 229/229 working source links; it resolved references to the first question after 16 unrelated questions. An answer cost 14,716 tokens (US\$0.06) with a median latency of 24 s.  Large-scale retrieval with agentic evidence selection and mechanical verification yields auditable, low-hallucination nutrition answers.
\end{abstract}


\IEEEpeerreviewmaketitle

\section{Introduction}\label{sec:1}

\IEEEPARstart{D}{iet} is one of the largest modifiable determinants of health. The Global Burden of Disease study attributed 11 million deaths and 255 million disability-adjusted life-years in 2017 to dietary risk factors, led by high sodium intake and low intake of whole grains and fruit~\cite{gbd2019}. Preventing and managing diet-related non-communicable disease therefore depends on people, clinicians and health services being able to obtain accurate, current and specific nutrition information. Much of that information is now sought online, where health misinformation is common.  A systematic review of 69 social-media studies reported misinformation rates of about 36\% for diet and eating-disorder topics~\cite{suarezlledo2021}. The health industry, from clinical nutrition services and digital-health providers to food and supplement information services, consequently needs tools that answer nutrition questions at scale while showing the evidence behind every statement.

Conversational large language models (LLMs) are a good  candidate and are already consulted for dietary advice. When dietitians and domain experts graded answers to eight common nutrition questions, ChatGPT's answers received higher overall grades than dietitians' answers for five questions and lower grades for none~\cite{kirk2023}, but expert panels found the appropriateness of ChatGPT's disease-specific dietary advice to range only from 55.5\% to 73.3\%, with some recommendations contradicting guidelines~\cite{ponzo2024}, and diet plans for people with food allergies contained unsafe items~\cite{niszczota2023}. More fundamentally, LLMs hallucinate~\cite{ji2023}; they fabricate a substantial share of the references they cite~\cite{walters2023}, and an automated audit of seven LLMs on 800 medical questions found that 50-90\% of responses were not fully supported by the sources they cited, with about 30\% of individual statements unsupported even for GPT-4o with web search~\cite{wu2025}. Specialized medical LLMs that pass examination questions~\cite{singhal2023} do not by themselves guarantee that a given statement is backed by a checkable source.

Retrieval-augmented generation (RAG) conditions the model on documents retrieved for each question~\cite{lewis2020,karpukhin2020,gao2023b}. In medicine, Almanac grounded clinical answers in curated resources and improved factuality, completeness and adversarial safety over general chatbots on 314 clinical questions~\cite{zakka2024}; Clinfo.ai answered clinical questions from dynamically retrieved PubMed literature~\cite{lozano2024}; and MedRAG with the MIRAGE benchmark (7,663 questions) showed that retrieval improves the accuracy of six LLMs by up to 18\% over chain-of-thought prompting~\cite{xiong2024}. Domain retrievers such as MedCPT improve biomedical search~\cite{jin2023}. Query-time strategies change how the model retrieves, IRCoT and Self-Ask interleave retrieval with reasoning~\cite{trivedi2023,press2023}, Self-RAG and Corrective RAG critique retrieved evidence~\cite{asai2024,yan2024}, deep-research agents plan and iterate searches~\cite{li2025,shao2024,nagori2025}, and biomedical variants add iterative follow-up questions~\cite{xiong2025}, self-critiqued PubMed queries~\cite{zhang2026}, self-reflection~\cite{ryan2026}, evidence verification~\cite{ning2025} and multi-source search~\cite{liu2025}.

Nutrition-specific RAG systems remain narrow in scope. A RAG-enhanced Llama~3 grounded in a 15,074-word American Heart Association knowledge bank outperformed off-the-shelf models on 30 cardiovascular nutrition questions rated by three experts~\cite{parameswaran2025}. iDISK2.0 answered dietary-supplement questions from a curated knowledge graph with 99\% and 95\% accuracy on true/false and multiple-choice questions~\cite{hou2025}. RAG improved the summarization of malnutrition information in aged-care health records~\cite{alkhalaf2024}, retrieval-augmented prompting was evaluated on 1,050 registered-dietitian examination questions~\cite{azimi2025}, and other systems target personalized food recommendation~\cite{yang2024}, meal-image feedback~\cite{jiang2025}, cardiac-health claims over PubMed~\cite{simons2025} and recipes~\cite{jeong2026}. A retrieval-focused comparison of twelve clinical RAG variants reported hallucination rates down to 5.8\% with self-reflective RAG~\cite{wolk2025}. Evaluation practice also differs, reference-free frameworks such as RAGAS score faithfulness and answer relevancy~\cite{es2024}, claim-level methods such as FActScore check atomic facts~\cite{min2023}, and citation quality can be measured separately from correctness~\cite{gao2023a}.

Several research gaps remain in the development and evaluation of retrieval-augmented generation (RAG) systems for health and nutrition applications. Existing systems primarily retrieve information from clinical guidelines, curated knowledge bases, or limited subsets of scientific literature, restricting their ability to access the broader scholarly record. Evaluation frameworks also remain fragmented, with most studies focusing on answer accuracy or expert assessments rather than jointly examining retrieval quality, response generation, evidence grounding, computational efficiency, latency, and cost. Furthermore, the reliability of cited evidence is rarely verified through automated checks of source existence, quotation accuracy, and numerical consistency. Another limitation is the reliance on isolated, self-contained questions for evaluation, despite real-world nutrition inquiries frequently involving multi-turn conversations and context-dependent follow-up questions. Additionally, nutrient quantities are often estimated by language models rather than retrieved directly from validated food composition databases, raising concerns about numerical accuracy and the reliability of dietary recommendations.

This study fills these gaps with SHAKTI (Smart Health Advisor for Knowledge-based Tailored Interventions). The proposed framework introduces several key innovations, including a comprehensive GPU-accelerated search index covering 484 million scholarly works from OpenAlex~\cite{priem2022}, complemented by a compressed index optimized for interactive retrieval. It enables a controlled comparison of four query-time RAG strategies using a common retrieval backend and language model to ensure consistent evaluation. To support multi-turn interactions, the framework incorporates a conversational memory mechanism that reformulates context-dependent queries into standalone questions while preserving relevant information from previous exchanges. Nutritional information is retrieved directly from USDA FoodData Central~\cite{fukagawa2022} to provide verified nutrient values rather than relying on language model estimates. Additionally, an automated evidence verification mechanism evaluates source accessibility, quotation accuracy, citation validity, and the consistency of numerical claims with their cited sources, presenting the results through a risk indicator. Finally, the framework employs a comprehensive evaluation approach covering retrieval performance, response generation, evidence grounding and factual accuracy, computational efficiency, and conversational context awareness. Claim-level verification is performed against the same retrieved source content available to the answering model, ensuring that factual assessments are grounded in the evidence used during response generation. The objective of this study is to design SHAKTI and to determine, on public benchmarks and in conversation, how accurately it retrieves evidence, how correct and how faithful to that evidence its answers are, how well it keeps the context of a conversation, and at what latency and cost.

The rest of the paper is organized as follows. Section~\ref{sec:2} describes the corpus, the questions, the retrieval models and RAG methods, SHAKTI's working mechanism and the evaluation design and metrics. Section~\ref{sec:3} reports the results for each metric family and for the conversational tests. Section~\ref{sec:4} compares them with health, diet and nutrition RAG systems. Section~\ref{sec:5} covers limitations and future work, and Section~\ref{sec:6} concludes the study.

\section{Methodology}\label{sec:2}

\subsection{Overview}\label{sec:2.1}

SHAKTI has an offline part, which embeds and indexes the scholarly corpus, and an online part, which answers each message in six stages (Fig.~\ref{fig:1}). The evaluation has three tiers: retrieval on 10,065 benchmark questions, a controlled comparison of four RAG methods on 1,000 of them, and tests of the deployed conversational system.

\subsection{Corpus and index}\label{sec:2.2}

The corpus is the OpenAlex snapshot of scholarly works~\cite{priem2022} (records updated up to 30 March 2026): 484,368,086 distinct works, of which 229,801,842 (47.4\%) have an abstract of at least 100 characters (the \textit{abstract-only} corpus). To characterize the corpus, works were grouped by their OpenAlex primary topic, subfield and field (one primary topic per work); 102.9 million works have no topic assigned. For each work we kept the identifier, title, abstract (reconstructed from the inverted index), year, DOI, venue and authors; works without an abstract were embedded from the title. Two bi-encoders were used to embed the works and the questions: MiniLM (all-MiniLM-L6-v2, 384 dimensions)~\cite{wang2020,reimers2019}, a general sentence encoder, and SPECTER2 (768 dimensions)~\cite{cohan2020,singh2023}, a scientific-document encoder with separate document and query adapters. The embeddings were stored as 16-bit (half-precision) floating-point vectors, which required 372 GB for MiniLM and 693 GB for SPECTER2. To confirm that no vectors had been corrupted during batch embedding and storage, randomly selected documents were embedded a second time and compared with their stored vectors; every pair had a cosine similarity of at least 0.98. For evaluation, every question was scored against every document on the GPU, keeping the exact top-100; one pass over the corpus took 29 min (MiniLM) and 56 min (SPECTER2), independent of the number of questions. Rankings were fused with reciprocal-rank fusion~\cite{cormack2009} and optionally reranked with the MedCPT cross-encoder~\cite{jin2023,nogueira2019}. For interactive use, the abstract-only corpus was compressed with product quantization~\cite{jegou2011,johnson2021} (48 bytes per paper, 11 GB on the GPU); its top-300 candidates contain 98.2\% of the exact MiniLM top-10.

\subsection{Questions and gold evidence}\label{sec:2.3}

We used four public benchmarks (Table~\ref{tab:1}): NFCorpus~\cite{boteva2016} (lay nutrition queries from NutritionFacts.org with the articles those pages cite, graded relevance), BioASQ~\cite{tsatsaronis2015} (biomedical questions with expert ideal answers), SciFact~\cite{wadden2020} (scientific claims with SUPPORT, CONTRADICT or NOT\_ENOUGH\_INFO verdicts) and PubMedQA~\cite{jin2019} (1,000 expert-labelled yes/no/maybe questions with long answers). Gold articles (the articles that a benchmark marks as relevant to a question) were resolved through NCBI E-utilities and matched to OpenAlex by DOI, or by normalized title when no DOI matched; any OpenAlex record of a gold article counts as retrieving it. Of 179,119 question-article links, 159,593 (89.1\%) were matched and 9,952 questions had at least one gold article in the corpus. A stratified random sample of 1,000 questions (250 per benchmark) was used for the RAG methods.

\begin{table*}[!t]
\centering
\caption{Benchmarks used as the question set. Gold links are question-article pairs; ``in corpus'' counts those matched to an OpenAlex work.}
\label{tab:1}
{\footnotesize\setlength{\tabcolsep}{4pt}
\begin{tabular}{@{}L{0.133\linewidth}L{0.113\linewidth}L{0.084\linewidth}L{0.118\linewidth}L{0.107\linewidth}L{0.351\linewidth}@{}}
\toprule
Benchmark & Questions & Scored & Gold links & In corpus & Reference answer \\
\midrule
NFCorpus & 3,237 & 3,197 & 134,294 & 115,192 & none (evidence only) \\
BioASQ & 4,719 & 4,718 & 42,608 & 42,267 & free text (expert ideal answer) \\
SciFact & 1,109 & 1,038 & 1,217 & 1,135 & SUPPORT / CONTRADICT / NEI \\
PubMedQA & 1,000 & 999 & 1,000 & 999 & yes / no / maybe + long answer \\
Total & 10,065 & 9,952 & 179,119 & 159,593 &  \\
\bottomrule
\end{tabular}}
\end{table*}

\subsection{Query-time RAG methods}\label{sec:2.4}

All methods used the same LLM, Claude Sonnet 5.5, called without tools and with JSON-schema-constrained output. They also shared one answering prompt, which instructs the model to use only the numbered papers, cite them by number, copy one exact proof sentence per key claim, and choose an insufficient-evidence verdict when the papers do not settle the question. Papers were given as title, year and the first 1,500 characters of the abstract. \textit{Single-shot RAG} answers from the ten highest-ranked papers of the hybrid retriever. \textit{Multi-hop RAG} (IRCoT/Self-Ask~\cite{trivedi2023,press2023}) starts from the top-5 and, up to twice, reasons about what is missing and issues one follow-up query (+5 papers). \textit{Corrective RAG} (CRAG~\cite{yan2024}) grades each of the top-10 papers as correct, ambiguous or incorrect, retrieves once more with a rewritten query if none is correct, and answers from the correct papers (topped up with ambiguous ones to three). The \textit{deep-research agent}~\cite{li2025,shao2024} plans three or four queries (direct question, trials or meta-analyses, mechanisms, contradicting evidence), reads up to 25 papers, keeps those with real evidence, issues up to two gap-filling queries and answers from up to 15 kept papers. Follow-up searches used the exact hybrid search over the abstract-only corpus.

\subsection{SHAKTI's working mechanism}\label{sec:2.5}

\begin{figure*}[!t]
\centering
\includegraphics[width=\textwidth,height=0.42\textheight,keepaspectratio]{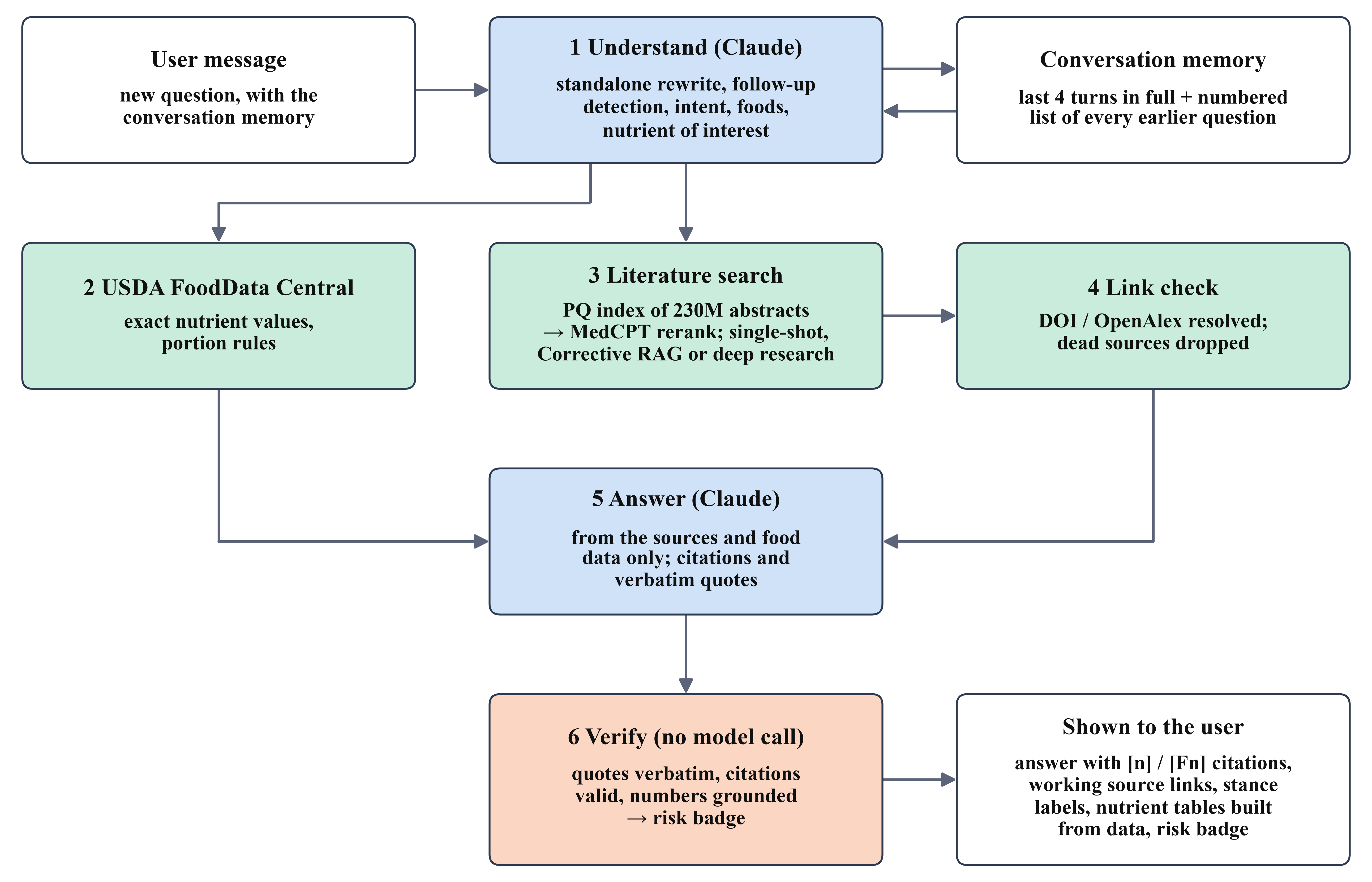}
\caption{SHAKTI's working mechanism for one chat message. Blue: Claude calls; green: data and retrieval; orange: online verification; white: the user's message, the conversation memory and what is shown to the user. The conversation memory feeds the understanding step; everything shown to the user is built from the sources or the food data.}
\label{fig:1}
\end{figure*}

Each chat message passes through six stages (Fig.~\ref{fig:1}).

The first stage interprets the message. Claude Sonnet receives the conversation memory and the new message and, under a JSON schema, returns a standalone rewrite of the message~\cite{elgohary2019}, states whether it is a follow-up and what it took from earlier turns, and identifies its intent (health evidence, food data, greeting, out of scope or emergency), the foods whose nutrients are needed, with their amounts and units, and the nutrient of interest. The memory holds the last four turns in full and a numbered list of every earlier question in its standalone form; because a session is capped at 20 questions, this list stays short. In the second stage, the foods are looked up in a local copy of USDA FoodData Central~\cite{fukagawa2022}, which supplies energy, macronutrients, fibre, sugars, sodium, saturated fat, cholesterol and twelve micronutrients (iron, vitamin C, calcium, potassium, magnesium, zinc, vitamins D, B12 and A, folate and caffeine); when no amount is given, a typical portion is chosen by deterministic rules that prefer standard measures and avoid unusual sizes.

The third stage retrieves the literature.  The standalone question is searched in the compressed index of 230 million abstracts (top-300 candidates), the candidates are reranked with MedCPT, and they are passed to one of three evidence strategies, single-shot, Corrective RAG or deep research, which are assigned at random and balanced per session for the planned user study. In the fourth stage, the DOI of every candidate paper (or, without a DOI, its OpenAlex record) is resolved, and papers whose records no longer exist are removed before answering. In the fifth stage, Claude Sonnet writes a plain-language answer only from the numbered papers and food values; it cites papers as [n] and foods as [Fn], copies verbatim proof sentences, labels the stance of each paper and lists numeric findings. In the sixth stage, without any model call, SHAKTI checks that every citation points to a provided source, that every quote occurs verbatim in the cited paper, that every number in the answer appears in the sources, the food data or the conversation (or is arithmetic on them), and that cited sentences are semantically close to a sentence of the cited paper (MiniLM cosine $\geq$ 0.45). The result is shown as a low, medium or high risk badge with reasons, and tables and charts are built from the food data and the verified sources, never drawn by the model.

All four checks run in a single deterministic function after the answer is generated, with no language model involved. Each check compares the answer against the evidence the generator was shown: the numbered papers (title and abstract), the matched food-composition rows, and the question plus the conversation so far. First, a regular expression extracts every citation group (e.g. [3], [2, 4], [1--3], [F1]). A citation is valid only if its number lies between 1 and the number of papers or foods provided; any other number is counted as fabricated. Second, each quote is lowercased, stripped of punctuation and whitespace-normalized, as is the title and abstract of the paper it cites, and it passes only if it appears there as an exact substring. Third, after citation brackets are removed and number words are converted to digits (``twenty-five'' $\rightarrow$ 25), every number above 3 in the answer must be grounded. Numbers 0--3 are skipped because they mostly mark list structure. A number is grounded if it appears in the papers' titles, abstracts or publication years, the question or the conversation, or if it matches a food-composition value (per 100 g, per serving, grams, or the sum over the matched foods) within a tolerance of $\pm$0.5 for values $\geq$ 10, $\pm$0.05 for smaller values, or 1\% of the value, whichever is largest. The sum or difference of two numbers already grounded also counts, so the answer's own arithmetic is accepted. Fourth, the answer is split into sentences, and each cited sentence is embedded with all-MiniLM-L6-v2 (L2-normalized) together with every sentence and every adjacent sentence pair of the cited papers' titles and abstracts. The sentence counts as supported if its highest cosine similarity to any of these passages is $\geq$ 0.45. These results set a risk level that is shown to the user with its reasons. Any invalid citation, fewer than half of the quotes verbatim, or two or more unverified numbers gives high risk. Weaker failures give medium risk, and the semantic check, being only a proxy for support, can raise the risk to medium but never to high. The checks therefore test whether the answer can be traced to its sources, not whether it is true, and factual accuracy was assessed separately by claim-level judging.

\subsection{Evaluation design}\label{sec:2.6}

\textit{Tier 1: retrieval.} All four retrievers (MiniLM, SPECTER2, hybrid, hybrid + MedCPT) on the full and abstract-only corpus, 9,952 scored questions. \textit{Tier 2: RAG methods.} The four methods on the 1,000-question sample. For faithfulness, hallucination and answer relevancy, a stratified subsample of 200 questions (50 per benchmark, the same for every method; 800 answers) was judged. \textit{Tier 3: SHAKTI.} (a) Twelve scripted conversations of three turns (36 turns, 21 follow-ups, 3 topic switches) on rice, fasting, coffee, nuts and vitamin D, ketogenic diets, eggs, green tea, breakfast, probiotics, cola and creatine, the Mediterranean diet and iron, each turn with a researcher-written reference standalone question and key concepts; run before and after adding micronutrients and portion rules. (b) A memory-depth test: a first question, \textit{N} unrelated questions (\textit{N} = 1, 3, 4, 5, 6, 8, 12, 16) and a call-back such as ``Going back to my very first question, how much fiber does that food have?'', two first questions per depth, before (last four turns only; 14 conversations, 106 questions) and after adding the earlier-question list (16 conversations, 142 questions). (c) A load test: 38 fresh questions from 1, 2, 4 and 8 concurrent virtual users.

\subsection{Evaluation Metrics}\label{sec:2.7}

Table~\ref{tab:2} defines the metrics for RAG.  \textit{Retrieval} is scored against the gold articles~\cite{jarvelin2002}. \textit{Generation} is scored against expert references, verdict exact match and macro-F1 for PubMedQA and SciFact; for the 500 sample questions with free-text references (BioASQ ideal answers, PubMedQA long answers) text exact match and token F1~\cite{rajpurkar2016}, ROUGE~\cite{lin2004}, BLEU~\cite{papineni2002} and BERTScore (roberta-large, baseline-rescaled)~\cite{zhang2020}, computed after removing citation markers; and, for BioASQ, an LLM grade of correctness against the ideal answer. \textit{Grounding and factuality} follows the claim-level approach of FActScore and RAGAS~\cite{es2024,min2023}. a Claude Opus judge (a stronger model than the answering model) splits each answer into atomic claims and labels each supported, partial, unsupported or contradicted by the sources, given exactly the text the system saw (same records, same 1,500-character cut; in conversation also the earlier answers); faithfulness is defined as (supported + $\frac{1}{2}$ partial)/claims and the hallucination rate is defined as (unsupported + contradicted)/claims. Context precision, context recall and citation accuracy use pooled TREC-style relevance judgments (0/1/2) of every evidence paper~\cite{zobel1998,thomas2024}, made by Claude Sonnet, together with the gold articles. \textit{Efficiency} uses the token usage and list-price cost recorded for every Claude call (replayed exactly from the call cache for Tier 2, logged per turn for Tier 3), latency and throughput. \textit{Context awareness} is summarized by a context-awareness score (CAS), the mean of follow-up detection accuracy, concept recall, rewrite fidelity, answer relevance on follow-ups and one minus the topic-leakage rate, each on a 0-1 scale (judge scores of 0-2 divided by 2) and reported as a percentage.

\begin{table*}[!t]
\centering
\caption{Metrics by family.}
\label{tab:2}
{\footnotesize\setlength{\tabcolsep}{4pt}
\begin{tabular}{@{}L{0.127\linewidth}L{0.842\linewidth}@{}}
\toprule
Family & Metrics (definition) \\
\midrule
Retrieval & Precision@10, Recall@10/100, Hit@10 (a gold article in the top 10), MRR@100, MAP@100 (mean of precision at each gold rank), nDCG@10 (graded for NFCorpus) \\
Generation & Verdict exact match and macro-F1; text exact match and token F1 (SQuAD normalization); ROUGE-1/2/L F-measure; corpus BLEU; BERTScore F1; BioASQ answer correct (LLM grade vs expert answer) \\
Grounding and factuality & Faithfulness; hallucination rate; contradiction rate; hallucination-free answers; answer relevancy (judge, 0-2 scaled to 0-1); context precision (rank-aware precision of relevant evidence); context recall (gold articles in the evidence); citation accuracy (cited papers that are direct evidence); proof validity (quote verbatim in a relevant cited paper); link validity \\
Efficiency & Claude calls, input and output tokens, list-price cost per answer; Claude time; latency p50/p95; throughput (answers per minute) \\
Context awareness & Follow-up detection accuracy and F1; concept recall of the rewrite (vs the raw message); rewrite fidelity (judge, 0-2); answer relevance on follow-ups; topic leakage; memory depth (call-backs resolved after \textit{N} unrelated questions) \\
\bottomrule
\end{tabular}}
\end{table*}

\subsection{Statistical analysis}\label{sec:2.8}

Means are reported with 95\% bootstrap confidence intervals~\cite{efron1993} (resampling questions, or conversations for Tier 3). Methods were compared with paired bootstrap tests on per-question differences~\cite{smucker2007} (two-sided; \textit{p} $<$ 10\textsuperscript{$-$4} reported as \textit{p} $<$ 0.001): hybrid and reranked retrieval against their baselines on 9,952 questions, and each RAG method against single-shot RAG on the 1,000-question sample. For zero events among \textit{n}  trials, we report the rule-of-three upper bound 3/\textit{n}.

\subsection{Implementation}\label{sec:2.9}

Embedding, search and reranking ran on one workstation (Intel Core i9-13900KF, 32 GB RAM, NVIDIA RTX 4080 16 GB); the GPU was used only for embedding, indexing, search and reranking. Claude was called through the Claude Code command-line interface with all responses cached on disk, so every result can be reproduced from the cache. SHAKTI runs as a local web server that answers at most two messages at a time and queues up to twelve.

\section{Results}\label{sec:3}

\subsection{Retrieval}\label{sec:3.1}

\begin{figure*}[!t]
\centering
\includegraphics[width=\textwidth,height=0.42\textheight,keepaspectratio]{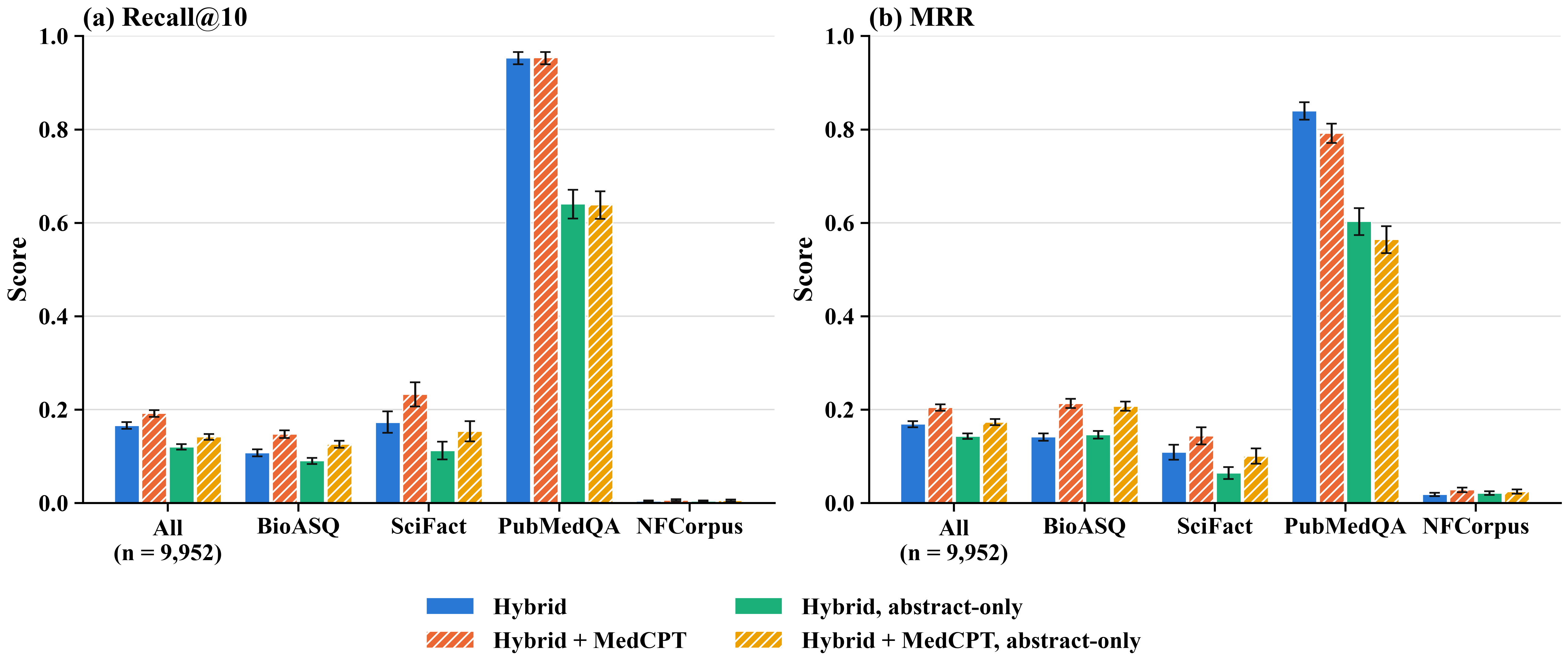}
\caption{Recall@10 and MRR of the hybrid retriever with and without MedCPT reranking, on the full and abstract-only corpus, overall and per benchmark (mean and 95\% bootstrap CI).}
\label{fig:2}
\end{figure*}

Table~\ref{tab:3} and Fig.~\ref{fig:2} report retrieval against the gold lists for all 9,952 scored questions. Hybrid fusion outperformed both MiniLM and SPECTER2 (Recall@10 0.17 vs 0.15 and 0.15; \textit{p} $<$ 0.001), and MedCPT reranking of the hybrid top-100 raised Recall@10 to 0.19, MRR to 0.21, MAP@100 from 0.13 to 0.14, nDCG@10 to 0.17 and Precision@10 from 0.03 to 0.04 (all \textit{p} $<$ 0.001). Gains were largest on BioASQ (Recall@10 0.11 to 0.15) and SciFact (0.17 to 0.23). Precision@10 is bounded by the small number of gold articles per question (most have one to three). Restricting the corpus to works with an abstract lowered gold-based scores (hybrid Recall@10 0.12), because a third of the gold articles have no abstract in OpenAlex.

Gold lists under-count relevant evidence at this scale. In the pooled judgments, 57.6\% of single-shot top-10 papers were relevant and 22.8\% direct evidence, while only 27.3\% of questions had a gold article in the top 10; for NFCorpus, 47.2\% of top-10 papers were relevant but the gold hit rate was 4.4\%. Gold-based scores therefore compare retrievers fairly but understate the share of useful evidence found. Low absolute scores are also expected because the gold lists were built for small collections (3,633 documents for NFCorpus and 5,183 for SciFact~\cite{thakur2021}), whereas here the same articles compete with 230 million (abstract-only corpus) to 484 million works (full corpus).

\begin{table*}[!t]
\centering
\caption{Retrieval against benchmark gold lists (9,952 questions). P@10: Precision@10; R@\textit{k}: Recall@\textit{k}; MRR: MRR@100; MAP: MAP@100; nDCG: nDCG@10. Best per column and corpus in bold.}
\label{tab:3}
{\footnotesize\setlength{\tabcolsep}{4pt}
\begin{tabular}{@{}L{0.138\linewidth}L{0.183\linewidth}C{0.074\linewidth}C{0.076\linewidth}C{0.087\linewidth}C{0.089\linewidth}C{0.069\linewidth}C{0.069\linewidth}C{0.076\linewidth}@{}}
\toprule
Corpus & Retriever & P@10 & R@10 & R@100 & Hit@10 & MRR & MAP & nDCG \\
\midrule
Full & MiniLM & 0.03 & 0.15 & 0.22 & 0.23 & 0.15 & 0.12 & 0.13 \\
 & SPECTER2 & 0.03 & 0.15 & 0.21 & 0.22 & 0.16 & 0.12 & 0.14 \\
 & Hybrid (RRF) & 0.03 & 0.17 & \textbf{0.23} & 0.25 & 0.17 & 0.13 & 0.15 \\
 & Hybrid + MedCPT & \textbf{0.04} & \textbf{0.19} & \textbf{0.23} & \textbf{0.32} & \textbf{0.21} & \textbf{0.14} & \textbf{0.17} \\
Abstract-only & MiniLM & 0.03 & 0.11 & 0.17 & 0.21 & 0.14 & 0.09 & 0.10 \\
 & SPECTER2 & 0.02 & 0.10 & 0.15 & 0.16 & 0.11 & 0.09 & 0.09 \\
 & Hybrid (RRF) & 0.03 & 0.12 & \textbf{0.18} & 0.22 & 0.14 & 0.10 & 0.11 \\
 & Hybrid + MedCPT & \textbf{0.04} & \textbf{0.14} & \textbf{0.18} & \textbf{0.28} & \textbf{0.17} & \textbf{0.11} & \textbf{0.13} \\
\bottomrule
\end{tabular}}
\end{table*}

\subsection{Generation}\label{sec:3.2}

Table~\ref{tab:4} reports answer quality against the expert references. Verdict accuracy (exact match of the predicted verdict) ranged from 0.61 to 0.66 on PubMedQA and from 0.56 to 0.58 on SciFact; Corrective RAG was highest on both. On BioASQ the deep-research agent answered 73.2\% of questions correctly against 59.2\% for single-shot RAG (\textit{p} $<$ 0.001). Text exact match, which credits an answer only when, after normalization, it reproduces the reference answer word for word, was 0.00 for every method and is therefore omitted from Table~\ref{tab:4}. This is expected rather than a failure: the references are short expert-written sentences or paragraphs, whereas the answers are longer explanations that add citations, verbatim quotes and numerical findings, so a correct answer never repeats the reference wording exactly. Overlap metrics were moderate and similar across methods (token F1 0.26-0.27, ROUGE-L 0.20-0.22, BLEU 0.05-0.06, BERTScore 0.20-0.22), while the judged correctness differed by 14 points; overlap metrics thus do not separate the methods that the correctness grade separates.

\begin{table*}[!t]
\centering
\caption{Generation metrics of the four RAG methods. Verdicts: 250 PubMedQA and 250 SciFact questions; BioASQ correct: 250; text metrics: 500 answers with an expert reference (250 BioASQ ideal answers, 250 PubMedQA long answers). Acc.: verdict accuracy (exact match of the predicted verdict); mF1: macro-F1; Corr.: BioASQ answer correct; Tok. F1: token F1; R-1, R-2, R-L: ROUGE-1, ROUGE-2, ROUGE-L; BERTS.: BERTScore F1, baseline-rescaled. Text exact match was 0.00 for every method and is not shown. The best value in each column is in bold (ties included). 95\% CIs in Fig.~\ref{fig:4}.}
\label{tab:4}
{\footnotesize\setlength{\tabcolsep}{4pt}
\begin{tabular}{@{}L{0.100\linewidth}C{0.056\linewidth}C{0.056\linewidth}C{0.056\linewidth}C{0.056\linewidth}C{0.087\linewidth}C{0.084\linewidth}C{0.054\linewidth}C{0.054\linewidth}C{0.054\linewidth}C{0.071\linewidth}C{0.087\linewidth}@{}}
\toprule
 & \multicolumn{2}{C{0.127\linewidth}}{PubMedQA} & \multicolumn{2}{C{0.127\linewidth}}{SciFact} & BioASQ & \multicolumn{6}{C{0.481\linewidth}}{Free-text answers vs reference (500)} \\
\cmidrule(lr){2-3}\cmidrule(lr){4-5}\cmidrule(lr){7-12}
Method & Acc. & mF1 & Acc. & mF1 & Corr. & Tok. F1 & R-1 & R-2 & R-L & BLEU & BERTS. \\
\midrule
Single-shot RAG & 0.61 & 0.57 & 0.57 & 0.57 & 0.59 & 0.26 & \textbf{0.30} & 0.11 & 0.21 & \textbf{0.06} & 0.20 \\
Multi-hop RAG & 0.64 & 0.59 & 0.56 & 0.56 & 0.63 & \textbf{0.27} & \textbf{0.30} & \textbf{0.12} & \textbf{0.22} & \textbf{0.06} & \textbf{0.22} \\
Corrective RAG & \textbf{0.66} & \textbf{0.60} & \textbf{0.58} & \textbf{0.58} & 0.66 & \textbf{0.27} & \textbf{0.30} & 0.11 & 0.21 & \textbf{0.06} & \textbf{0.22} \\
Deep-research agent & 0.64 & 0.57 & 0.57 & 0.56 & \textbf{0.73} & 0.26 & 0.29 & 0.11 & 0.20 & 0.05 & 0.20 \\
\bottomrule
\end{tabular}}
\end{table*}

\subsection{Grounding and factuality}\label{sec:3.3}

Table~\ref{tab:5} and Figs.~3 and 4 report grounding. All methods were highly faithful to their evidence: faithfulness 0.95-0.96 and hallucination rates of 0.5-0.8\% of claims (6,401 claims in 800 answers; 0.1\% or fewer contradicted), with 94.0-95.5\% of answers free of any unsupported claim. The methods differed mainly in what evidence they used. Corrective RAG had the highest context precision (0.98 vs 0.73 for single-shot RAG) because it discards papers it grades as irrelevant; the deep-research agent had the highest context recall (0.25), citation accuracy (0.72), proof validity (0.97) and answer relevancy (0.98 vs 0.89). Every multi-step method improved every evidence metric against single-shot RAG (\textit{p} $<$ 0.001; Table~\ref{tab:6}).

\begin{table*}[!t]
\centering
\caption{Grounding and factuality of the four RAG methods. Context precision, context recall, citation accuracy and proof validity: 1,000 questions; faithfulness, hallucination, hallucination-free answers and answer relevancy: 200 questions per method, a stratified subsample of the 1,000 (50 per benchmark, the same questions for every method). The first four columns therefore describe all 1,000 answers of each method, and the last four the 200 of these answers that were judged claim by claim  (Claude Opus claim judge; 95\% CI in brackets).}
\label{tab:5}
{\footnotesize\setlength{\tabcolsep}{4pt}
\begin{tabular}{@{}L{0.111\linewidth}C{0.077\linewidth}C{0.085\linewidth}C{0.085\linewidth}C{0.077\linewidth}C{0.124\linewidth}C{0.098\linewidth}C{0.077\linewidth}C{0.126\linewidth}@{}}
\toprule
Method & Context prec. & Context recall & Citation acc. & Proof valid & Faithfulness & Hallucination & Halluc.-free & Answer relevancy \\
\midrule
Single-shot RAG & 0.73 & 0.22 & 0.54 & 0.89 & \textbf{0.96}   \newline [0.95, 0.97] & 0.8\%  \newline [0.3, 1.5] & \textbf{0.96} & 0.89  \newline [0.86, 0.92] \\
Multi-hop RAG & 0.79 & 0.21 & 0.66 & 0.96 & 0.95  \newline [0.94, 0.96] & 0.8\%  \newline [0.4, 1.3] & 0.94 & 0.95  \newline [0.93, 0.97] \\
Corrective RAG & \textbf{0.98} & 0.22 & 0.66 & 0.95 & 0.96  \newline [0.95, 0.97] & 0.8\%  \newline [0.4, 1.4] & 0.95 & 0.93  \newline [0.91, 0.95] \\
Deep-research agent & 0.96 & \textbf{0.25} & \textbf{0.72} & \textbf{0.97} & 0.96  \newline [0.95, 0.96] & \textbf{0.5\%}   \newline [0.2, 0.8] & 0.95 & \textbf{0.98}   \newline [0.97, 0.99] \\
\bottomrule
\end{tabular}}
\end{table*}

\begin{table*}[!t]
\centering
\caption{Paired differences against single-shot RAG (1,000 questions; mean difference, 95\% bootstrap CI). Verdict accuracy pools PubMedQA and SciFact. \textsuperscript{*}\textit{p} $<$ 0.05; \textsuperscript{**}\textit{p} $<$ 0.01; \textsuperscript{***}\textit{p} $<$ 0.001.}
\label{tab:6}
{\footnotesize\setlength{\tabcolsep}{4pt}
\begin{tabular}{@{}L{0.233\linewidth}C{0.235\linewidth}C{0.235\linewidth}C{0.235\linewidth}@{}}
\toprule
Metric & Multi-hop & Corrective & Deep research \\
\midrule
Verdict accuracy & +0.01 ($-$0.02, 0.04) & +0.03\textsuperscript{*} (0.00, 0.05) & +0.01 ($-$0.03, 0.05) \\
BioASQ answer correct & +0.04 ($-$0.01, 0.08) & +0.06\textsuperscript{**} (0.02, 0.10) & +0.14\textsuperscript{***} (0.09, 0.20) \\
Judged precision & +0.13\textsuperscript{***} (0.12, 0.14) & +0.36\textsuperscript{***} (0.34, 0.37) & +0.28\textsuperscript{***} (0.26, 0.30) \\
Direct-evidence rate & +0.17\textsuperscript{***} (0.14, 0.19) & +0.11\textsuperscript{***} (0.09, 0.13) & +0.24\textsuperscript{***} (0.21, 0.27) \\
Citation accuracy & +0.12\textsuperscript{***} (0.10, 0.14) & +0.12\textsuperscript{***} (0.11, 0.14) & +0.18\textsuperscript{***} (0.16, 0.20) \\
Proof validity & +0.07\textsuperscript{***} (0.05, 0.09) & +0.06\textsuperscript{***} (0.05, 0.08) & +0.08\textsuperscript{***} (0.07, 0.10) \\
Gold hit & 0.00 ($-$0.02, 0.01) & +0.01 (0.00, 0.02) & +0.05\textsuperscript{***} (0.04, 0.07) \\
\bottomrule
\end{tabular}}
\end{table*}

\begin{figure*}[!t]
\centering
\includegraphics[width=\textwidth,height=0.42\textheight,keepaspectratio]{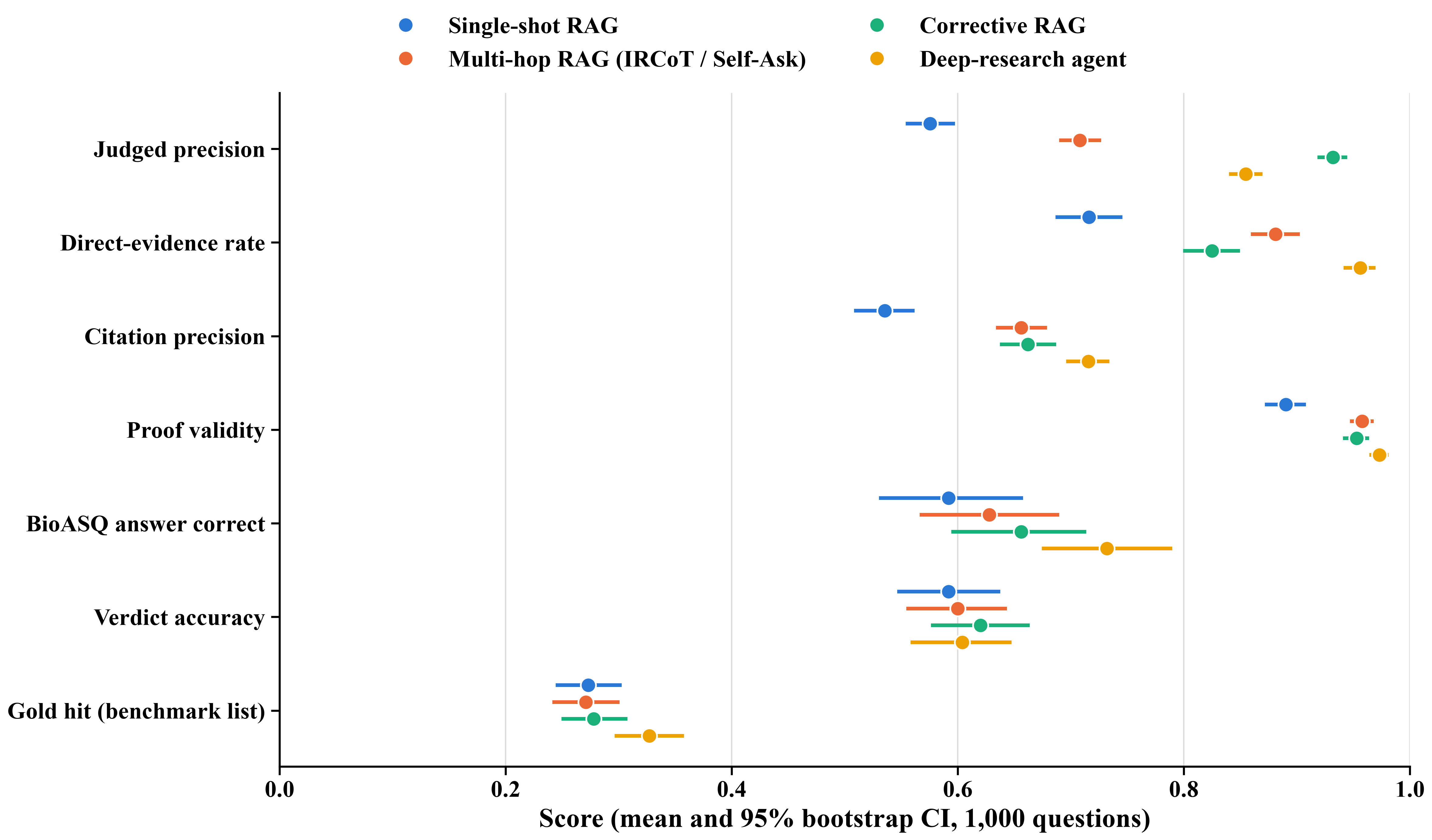}
\caption{Evidence metrics of the four RAG methods (mean and 95\% bootstrap CI over 1,000 questions; verdict accuracy over 500 and BioASQ answer correctness over 250).}
\label{fig:3}
\end{figure*}

\begin{figure*}[!t]
\centering
\includegraphics[width=\textwidth,height=0.42\textheight,keepaspectratio]{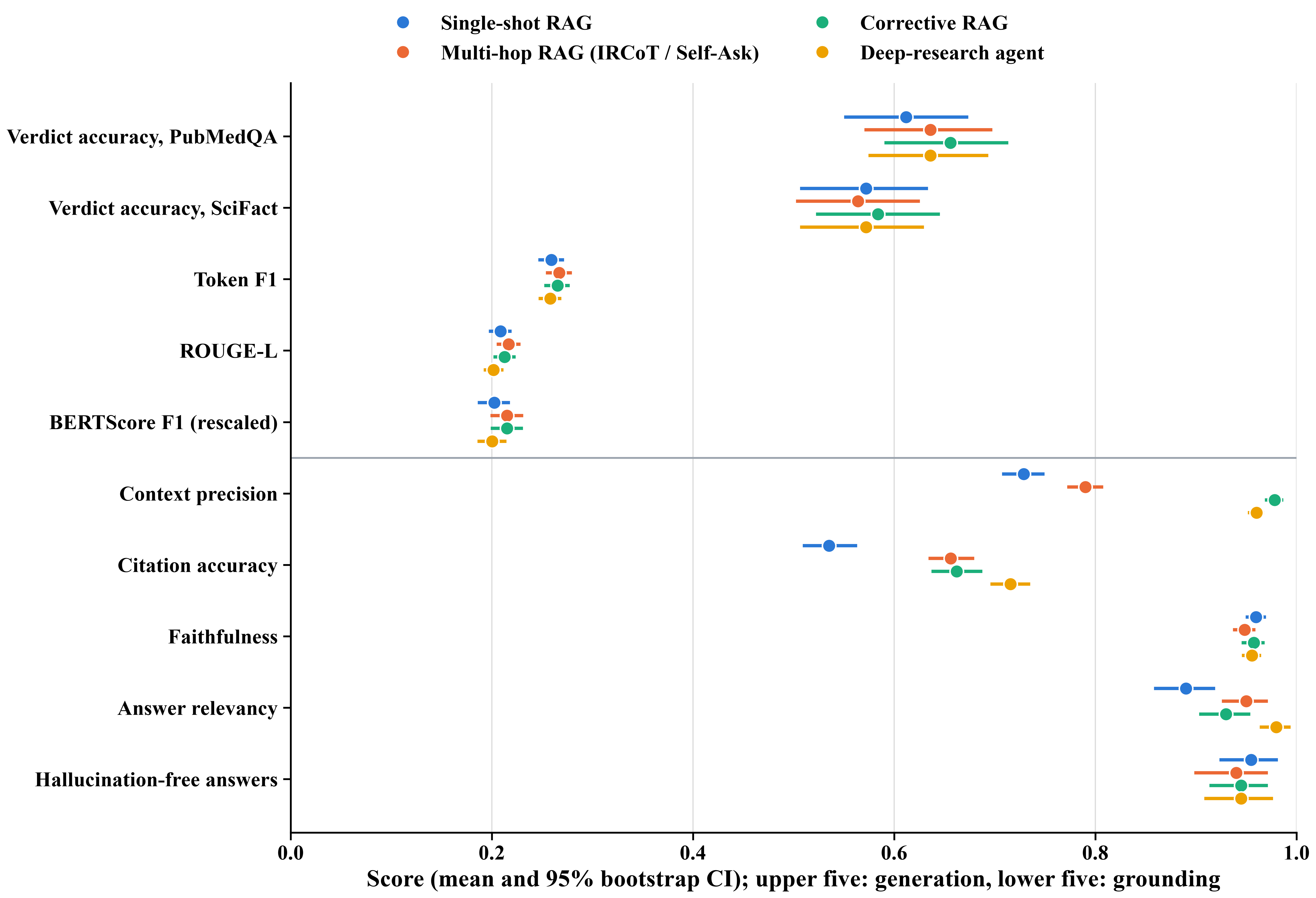}
\caption{Generation (top) and grounding (bottom) metrics of the four RAG methods (mean and 95\% bootstrap CI). Overlap metrics are low and flat across methods, whereas grounding is uniformly high and evidence selection differs.}
\label{fig:4}
\end{figure*}

\subsection{SHAKTI in conversation}\label{sec:3.4}

\begin{figure*}[!t]
\centering
\includegraphics[width=\textwidth,height=0.42\textheight,keepaspectratio]{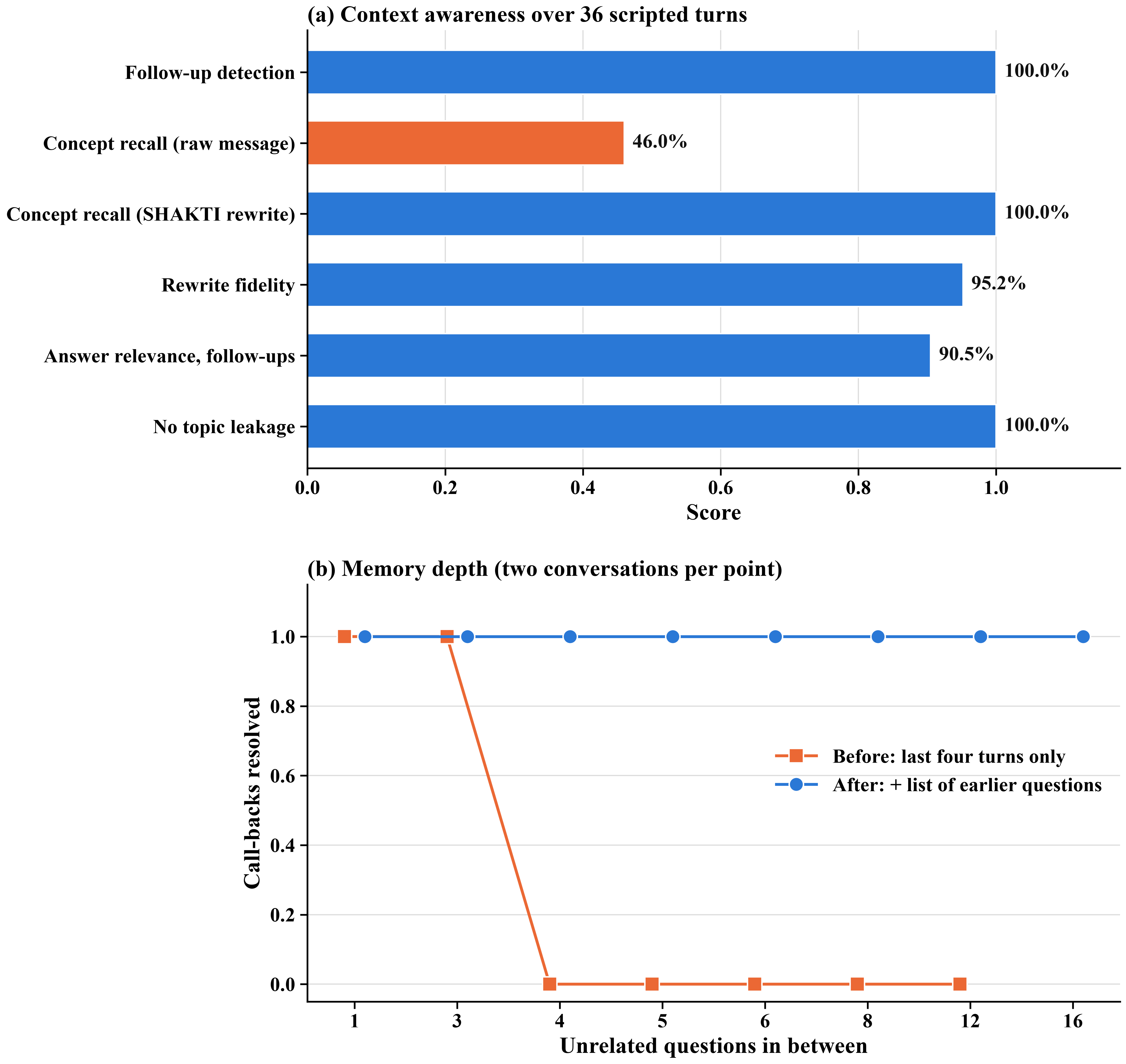}
\caption{(a) Context-awareness components over 36 scripted turns; concept recall of the raw message (orange) is the no-context baseline. (b) Share of call-backs to the first question resolved after \textit{N} unrelated questions, with the last four turns as memory (orange) and with the added list of earlier questions (blue).}
\label{fig:5}
\end{figure*}

Table~\ref{tab:7} and Fig.~\ref{fig:5} summarize the 36 scripted turns. SHAKTI detected every follow-up and every topic switch (follow-up F1 1.00) and its standalone rewrites contained all key concepts of every follow-up (concept recall 100\%, against 46.0\% in the raw messages), with a rewrite fidelity of 95.2\%; no topic leaked into the next conversation topic (0/3; rule-of-three bound 100\%, a small sample). The context-awareness score was 97.1\% (the mean of 100\%, 100\%, 95.2\%, 90.5\% and 100\%). The judge split the 36 answers into 377 claims: 350 supported, 20 partial, 7 unsupported and none contradicted, giving faithfulness 95.5\% and a hallucination rate of 1.9\% (0.3\% excluding general-knowledge statements); 80.6\% of answers were free of unsupported claims. Online checks found every quote verbatim, no citation pointing to a missing source and 98.4\% of numbers traced to the sources or food data; all 229 source links resolved, after 3 dead records had been removed before answering. The risk badge was low for 33 answers, medium for 2 and high for 1. Before micronutrients and portion rules were added, answers on the food-data path addressed the question in 81.8\% of cases (11 turns); afterwards in 100\%, and overall answer relevance rose from 90.3\% to 94.4\% while the hallucination rate stayed at 1.8-1.9\%.

\begin{table*}[!t]
\centering
\caption{SHAKTI in 36 scripted conversational turns (12 conversations; 21 follow-ups; Claude Opus judge for claims, rewrite fidelity and relevance; 95\% CI by resampling conversations).}
\label{tab:7}
{\footnotesize\setlength{\tabcolsep}{4pt}
\begin{tabular}{@{}L{0.466\linewidth}L{0.255\linewidth}L{0.232\linewidth}@{}}
\toprule
Metric & Result & Scored on \\
\midrule
Context-awareness score & 97.1\% & 36 turns \\
Follow-up detection (accuracy; F1) & 100\%; 1.00 & 36 turns \\
Concept recall: raw message $\rightarrow$ SHAKTI rewrite & 46.0\% $\rightarrow$ 100\% & 21 follow-ups \\
Rewrite fidelity & 95.2\% [88.6, 100] & 21 follow-ups \\
Answer relevance (all; follow-ups) & 94.4\% [87.5, 100]; 90.5\% & 36; 21 turns \\
Topic leakage & 0/3 & 3 topic switches \\
Faithfulness & 95.5\% [92.8, 97.6] & 377 claims, 36 answers \\
Hallucination rate (unsupported + contradicted) & 1.9\% [0.5, 3.4] & 377 claims \\
Contradicted claims & 0 & 377 claims \\
Hallucination-free answers & 80.6\% & 36 answers \\
Claim citation precision & 97.8\% & cited claims \\
Quotes verbatim; invalid citations & 100\%; 0 & 25; 36 answers \\
Numbers grounded in sources or food data & 98.4\% & 35 answers \\
Working source links & 229/229 & 229 links \\
\bottomrule
\end{tabular}}
\end{table*}

In the memory-depth test, \textit{w}ith only the last four turns as memory, SHAKTI resolved a call-back by the user to the first question after 1 or 3 unrelated questions (4/4) but never after 4 or more (0/10): it attached ``my first question'' to the oldest question still visible. With the numbered list of earlier questions added, it resolved all 16 call-backs at every depth up to 16 unrelated questions, and all 16 answers addressed the original topic (Fig.~\ref{fig:5}b).

\subsection{Efficiency}\label{sec:3.5}

\begin{figure*}[!t]
\centering
\includegraphics[width=\textwidth,height=0.42\textheight,keepaspectratio]{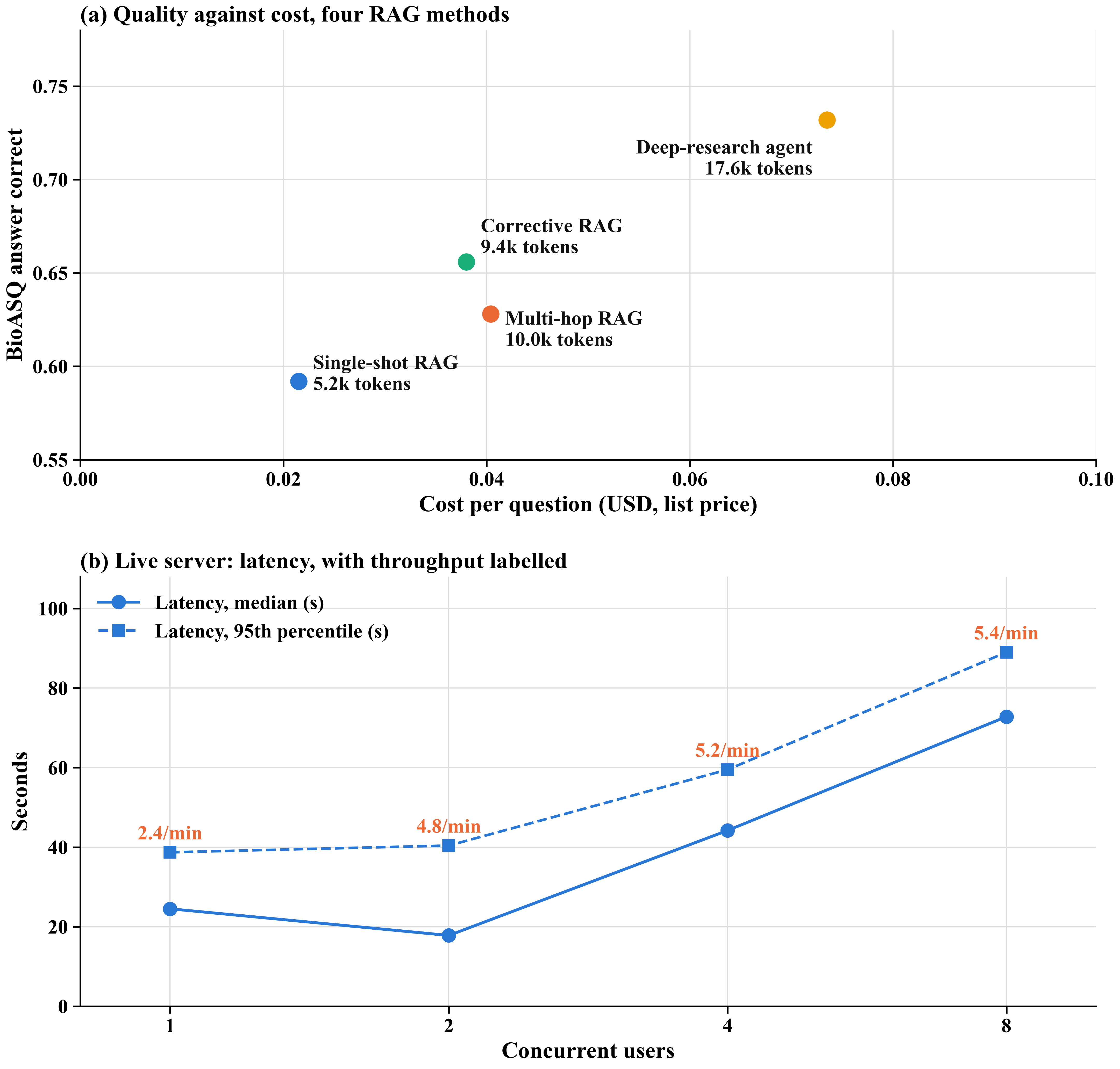}
\caption{(a) BioASQ answer correctness against cost per question for the four RAG methods (labels: tokens per question). (b) Live latency (median and 95th percentile) and throughput (labels) by number of concurrent users.}
\label{fig:6}
\end{figure*}

Table~\ref{tab:8} and Fig.~\ref{fig:6} report efficiency. Per question, single-shot RAG used one Claude call and 5,161 tokens (US\$0.02 at list price), and the deep-research agent three calls, 17,607 tokens and 5.3 extra searches (US\$0.07), about 3.4 times the cost for 14 more points of BioASQ correctness. Batch retrieval with both encoders over all 230 million abstracts took 0.50 s per question (120 questions per minute). In the live system, an answer used 2.8 Claude calls and 14,716 tokens on average (US\$0.06; US\$58 per 1,000 answers), from 6,356 tokens on the food-data path to 26,269 for deep research. With one user, the median latency was 24.5 s (95th percentile 38.7 s), split mainly between writing the answer (10.3 s), searching (6.5 s) and understanding the question (4.1 s), while link checking and verification took under 0.5 s. Throughput rose from 2.4 answers per minute with one user to 5.4 with eight, where the server's limit of two concurrent answers made the median latency 72.8 s, mostly queueing (54.5 s); all 38 questions were answered without error.

\begin{table*}[!t]
\centering
\caption{Efficiency. Top: four RAG methods (999 questions each, every Claude call replayed from the cache with its recorded usage). Bottom: the live SHAKTI server (38 fresh questions). Cost at Anthropic list prices.}
\label{tab:8}
{\footnotesize\setlength{\tabcolsep}{4pt}
\begin{tabular}{@{}L{0.158\linewidth}C{0.098\linewidth}C{0.099\linewidth}C{0.110\linewidth}C{0.099\linewidth}C{0.117\linewidth}C{0.097\linewidth}C{0.097\linewidth}@{}}
\toprule
Method & Claude calls & Input tokens & Output tokens & Total tokens & Cost / question & Cost / 1,000 & Claude time \\
\midrule
Single-shot RAG & 1.00 & 4,596 & 565 & 5,161 & \$0.02 & \$21.5 & 7.2 s \\
Multi-hop RAG & 2.39 & 8,965 & 1,000 & 9,965 & \$0.04 & \$40.4 & 15.7 s \\
Corrective RAG & 2.18 & 8,504 & 917 & 9,422 & \$0.04 & \$38.0 & 13.1 s \\
Deep-research agent & 3.00 & 16,068 & 1,539 & 17,607 & \$0.07 & \$73.5 & 20.2 s \\
SHAKTI live (mean) & 2.82 & 12,828 & 1,888 & 14,716 & \$0.06 & \$58.0 & 18.5 s \\
\bottomrule
\end{tabular}}
\par\medskip
{\footnotesize\setlength{\tabcolsep}{4pt}
\begin{tabular}{@{}L{0.137\linewidth}C{0.088\linewidth}C{0.205\linewidth}C{0.104\linewidth}C{0.104\linewidth}C{0.124\linewidth}C{0.128\linewidth}@{}}
\toprule
Concurrent users & Answered & Throughput (answers/min) & Latency p50 & Latency p95 & Processing p50 & Queue wait p50 \\
\midrule
1 & 4/4 & 2.4 & 24.5 s & 38.7 s & 24.1 s & 0 s \\
2 & 6/6 & 4.8 & 17.8 s & 40.4 s & 17.0 s & 0 s \\
4 & 12/12 & 5.2 & 44.2 s & 59.5 s & 18.5 s & 18.2 s \\
8 & 16/16 & 5.4 & 72.8 s & 89.0 s & 20.1 s & 54.5 s \\
\bottomrule
\end{tabular}}
\end{table*}

\section{Discussion}\label{sec:4}

SHAKTI answered health and nutrition questions from the full scholarly record with consistently high grounding: in every configuration fewer than 2\% of claims were unsupported by the given sources and almost none were contradicted, every live source link worked and every live quote was verbatim. Three further findings stand out. Agentic retrieval mainly improves \textit{which} evidence is used (context precision, citation accuracy, proof validity) and, for open questions, correctness, while closed-form verdicts barely change. Reference-overlap metrics hardly distinguish methods whose judged correctness differs by 14 points. And a conversational assistant needs an explicit memory of earlier questions; a fixed window of recent turns silently loses the thread.

\subsection{Comparison with health, diet and nutrition RAG studies}\label{sec:4.1}

Table~\ref{tab:9} sets SHAKTI against related systems. The studies differ in task, knowledge source, model and judge, so the comparison is of reported figures, not a head-to-head experiment.

\begin{table*}[!t]
\centering
\caption{Comparison with health, diet and nutrition RAG and LLM studies (figures as reported in each publication).}
\label{tab:9}
{\footnotesize\setlength{\tabcolsep}{4pt}
\begin{tabular}{@{}L{0.152\linewidth}L{0.153\linewidth}L{0.169\linewidth}L{0.236\linewidth}L{0.212\linewidth}@{}}
\toprule
Study & Domain; knowledge source & Evaluation size & Reported result & This study (SHAKTI) \\
\midrule
Parameswaran et al.~\cite{parameswaran2025} & CVD nutrition; 15,074-word AHA knowledge bank & 30 questions $\times$ 3, 3 experts & RAG Llama 3 rated best on reliability, appropriateness, adherence; no harm & 484M works + USDA data; 10,065 / 1,000 / 36 questions; claim-level faithfulness 95.5-96.0\% \\
Hou et al.~\cite{hou2025} (iDISK2.0) & Dietary supplements; curated knowledge graph & 1,000 true/false + 1,000 MCQ & 99\% and 95\% accuracy (GPT-4o alone 62\%, 52\%) & 3-class research verdicts from open retrieval: 61.2-65.6\% (PubMedQA), 56.4-58.4\% (SciFact) \\
Xiong et al.~\cite{xiong2024} (MedRAG) & Medical QA; PubMed, textbooks, other corpora & 7,663 questions & Accuracy up to +18\% over chain-of-thought & Agentic RAG: BioASQ correct +14.0 points over single-shot RAG; verdicts +0.8 to +2.8 \\
Wu et al.~\cite{wu2025} (SourceCheckup) & Medical queries; LLMs' own citations & 800 questions, 58,000 statement-source pairs & 50-90\% of responses not fully supported; GPT-4o with web search $\approx$30\% of statements unsupported & 0.5-1.9\% of claims unsupported; 80.6-95.5\% of answers fully supported \\
Wołk~\cite{wolk2025} & Clinical decision support; 12 RAG variants & 250 vignettes & Self-reflective RAG hallucination 5.8\%; hybrid RRF P@5 $\geq$ 0.68 & Claim hallucination 0.5-1.9\%; retrieval at 484M-work scale \\
Zakka et al.~\cite{zakka2024} (Almanac) & Clinical guidelines; curated resources & 314 questions, 10 raters & Better factuality, completeness, preference and adversarial safety than ChatGPT-4, Bing, Bard & Factuality measured automatically per claim; 0 contradicted claims in conversation \\
Alkhalaf et al.~\cite{alkhalaf2024} & Malnutrition in aged-care EHRs & 25 + 1,399 notes & Summary accuracy 93.25\% $\rightarrow$ 99.25\% with RAG & Open-domain questions rather than summarization of given records \\
Jeong and Suh~\cite{jeong2026} & Korean recipes & 60 queries, 180 responses & RAGAS faithfulness and relevancy not correlated with user ratings & User study planned (Section~\ref{sec:5}) \\
Ponzo et al.~\cite{ponzo2024} & NCD dietary advice; ChatGPT without retrieval & disease scenarios, expert panel & Appropriateness 55.5-73.3\%; two recommendations contradicted guidelines & No contradicted claims in 377 conversational claims \\
\bottomrule
\end{tabular}}
\end{table*}

SHAKTI's strongest results are in grounding, where it compares favorably with every figure we found. SourceCheckup reported that 50-90\% of LLM responses to medical queries were not fully supported by their cited sources and that even GPT-4o with web search left about 30\% of statements unsupported~\cite{wu2025}; in SHAKTI 0.5-1.9\% of claims were unsupported and 80.6-95.5\% of answers were fully supported. Self-reflective RAG had the lowest hallucination rate among twelve clinical RAG variants, 5.8\%~\cite{wolk2025}, against 0.5-0.8\% for our four methods. Unlike reports of fabricated references~\cite{walters2023}, no SHAKTI citation pointed to a missing source and all 229 links resolved, because links are checked before answering and citations and quotes are verified after it. The comparison is not like-for-like, for two reasons. SourceCheckup checks support by the cited source only, whereas our hallucination rate counts a claim as supported if any provided source supports it (claim citation precision, which checks the cited source, was 97.8\% in conversation and 98.4-99.7\% for the methods), and the studies used different judges and question sets. Compared with general chatbots giving dietary advice, where some recommendations contradicted guidelines~\cite{ponzo2024}, SHAKTI produced no contradicted claim in conversation and at most 0.1\% in the benchmark answers.

On closed-form questions SHAKTI does not outperform specialized systems. iDISK2.0 reached 99\% and 95\% accuracy on true/false and multiple-choice supplement questions answered from a curated knowledge graph~\cite{hou2025}; our verdict accuracy on PubMedQA, a three-class task over open retrieval, was 61.2-65.6\%, above the 55.2\% majority baseline but below the 78.0\% of a single human reading the source abstract~\cite{jin2019}. Agentic retrieval raised BioASQ correctness by 14.0 points over single-shot RAG, in line with the gains that MedRAG reported for retrieval over chain-of-thought prompting~\cite{xiong2024} and that iterative follow-up questions gave in medical QA~\cite{xiong2025}, but it moved verdicts by at most 2.8 points. This agrees with the observation that retrieval-oriented gains overstate generation gains of advanced RAG~\cite{chen2026}: deciding between ``maybe'' and ``no'', or between NOT\_ENOUGH\_INFO and CONTRADICT, is limited by reasoning over the evidence rather than by finding it.

Table~\ref{tab:10} lists the systems that report higher scores than SHAKTI on the same benchmarks, together with the conditions under which each score was obtained. Only one comparison is direct. With the source abstract withheld and the evidence retrieved, as in SHAKTI, GPT-4 with MedRAG reached 70.6\% on PubMedQA*~\cite{xiong2024}, 5 to 9 points above SHAKTI's 61.2-65.6\%. This gap is close to the 95\% confidence interval of about $\pm$6 points that follows from our 250 PubMedQA questions, and MedRAG was not required to cite or quote its evidence, whereas SHAKTI had to quote a proof sentence and could answer that the evidence was insufficient. The other scores were obtained with the source abstract supplied~\cite{nori2023a,nori2023b,singhal2025}, on binary yes/no questions~\cite{xiong2024}, on question subsets filtered to be answerable from the retrieval corpus~\cite{stuhlmann2025}, within small, fully judged collections or with a different metric~\cite{thakur2021,wadden2022}, or without enough detail on the question set and setup to judge~\cite{kim2025,ryan2026}. They show what easier versions of these tasks allow rather than ranking the systems, and a like-for-like comparison would require running these methods on SHAKTI's question sample and corpus.

\begin{table*}[!t]
\centering
\caption{Systems reporting higher scores than SHAKTI on the same benchmarks, and whether the comparison is direct (figures as reported in each publication; SHAKTI: 250 questions per benchmark, open retrieval over 484 million works, cited and quoted answers).}
\label{tab:10}
{\footnotesize\setlength{\tabcolsep}{4pt}
\begin{tabular}{@{}L{0.170\linewidth}L{0.131\linewidth}L{0.228\linewidth}L{0.102\linewidth}L{0.141\linewidth}L{0.137\linewidth}@{}}
\toprule
Study & Benchmark (questions) & Setup & Reported result & SHAKTI & Comparable? \\
\midrule
Xiong et al.~\cite{xiong2024}, GPT-4 + MedRAG & PubMedQA* (500) & Abstract withheld; retrieval from PubMed, textbooks, StatPearls and Wikipedia; no citation required & 70.6\% & 61.2-65.6\% & Yes, closest setup \\
Ryan et al.~\cite{ryan2026}, Self-MedRAG & PubMedQA (about 500 of 1,000 sampled with MedQA) & BM25 + Contriever retrieval; corpus not specified & 79.8\% & 61.2-65.6\% & Partly; corpus unclear \\
Nori et al.~\cite{nori2023b}, GPT-4 & PubMedQA reasoning-required (500) & Source abstract given; five-shot & 80.4\% & 61.2-65.6\% & No; no retrieval \\
Singhal et al.~\cite{singhal2025}, Med-PaLM 2 & PubMedQA reasoning-required (500) & Abstract given; model fine-tuned on benchmark subsets & 81.8\% & 61.2-65.6\% & No; no retrieval \\
Nori et al.~\cite{nori2023a}, GPT-4 + Medprompt & PubMedQA reasoning-required (500) & Abstract given; few-shot, chain-of-thought and ensembling & 82.0\% & 61.2-65.6\% & No; no retrieval \\
Kim~\cite{kim2025}, MedBioRAG & PubMedQA and BioASQ & Question sets and setup not specified; preprint & 85\%; 96\% & 61.2-65.6\%; 59.2-73.2\% & No; insufficient detail \\
Xiong et al.~\cite{xiong2024}, GPT-4 + MedRAG & BioASQ yes/no (618) & Two answer options & 92.6\% & 59.2-73.2\% (free-text answers) & No; binary task \\
Stuhlmann et al.~\cite{stuhlmann2025}, BM25 + MedCPT & BioASQ Task B & Factoid and list questions removed; only questions answerable from the corpus kept & 0.90 accuracy & 59.2-73.2\% & No; filtered questions \\
Wadden et al.~\cite{wadden2022}, MultiVerS & SciFact test claims & Closed SciFact abstract corpus; abstract-level label F1 & 0.73 & 0.56-0.58 (macro-F1, open retrieval) & No; closed corpus, different metric \\
Thakur et al.~\cite{thakur2021}, BEIR baselines & NFCorpus (323 queries over 3,633 documents, fully judged) & NDCG@10 & 0.33-0.35 & Recall reported over 484M works & No; different collection and metric \\
\bottomrule
\end{tabular}}
\end{table*}

Hybrid fusion and MedCPT reranking confirm, at a corpus an order of magnitude larger and multidisciplinary, the benefits reported for combining retrievers and using domain retrievers in MedRAG~\cite{xiong2024,jin2023}. Precision@5 of 0.68 or more in the clinical RAG comparison~\cite{wolk2025} and our Precision@10 of 0.04 are not comparable: their collection is small and fully judged, while our gold lists cover a few articles per question in 484 million works. Pooled judgments show that most retrieved papers are relevant, and gold lists built over small collections become incomplete at web scale~\cite{zobel1998,buckley2004,voorhees2020,reimers2021}.

Retrieval was not restricted to nutrition literature because nutrition evidence is spread across fields: only 7.9\% of NFCorpus's relevant papers are classified under Nutrition and Dietetics or Food Science, and 28.9\% under any nutrition-related topic (Supplementary Table~\ref{tab:S1}). A nutrition-only index (7.56 million works, about 1.5\% of the corpus) would have made 71\% of NFCorpus's relevant papers, and 78\% across all four benchmarks, unretrievable, and even a 114-million-work health, life-science and agriculture subset excludes 10\%; these are coverage ceilings (the share of relevant papers a slice contains), not measured changes in nDCG or recall, as no retrieval run was restricted to a slice. The full index is also inexpensive: batch retrieval over all 230 million abstracts took 0.50 s per question, and in the live system search took 6.5 s of the 24.5 s median answer time, so a smaller index could have shortened answers by at most about a quarter. Generation does not draw on the index: the answering model is not fine-tuned on it and answers only from the papers retrieved at question time.

Prior nutrition RAG studies evaluated 30 questions~\cite{parameswaran2025}, 2,000 closed-form questions~\cite{hou2025} or 1,050 examination questions~\cite{azimi2025}, over guideline text or knowledge graphs, and none evaluated multi-turn conversation. SHAKTI was evaluated on 10,065 retrieval questions, 1,000 end-to-end questions and conversational tests, over all 484 million works. Its query rewriting~\cite{elgohary2019} raised concept recall on follow-ups from 46.0\% to 100\%, and the earlier-question memory extended reliable call-backs from 3 to at least 16 intervening questions. Nutrient amounts are taken from FoodData Central~\cite{fukagawa2022} and checked online, which removes a known error source of LLM nutrition estimates.

Few health RAG studies report cost or latency; MedRAG reported the total prompt tokens of its experiments rather than per-answer costs~\cite{xiong2024}. SHAKTI's US\$0.06 per answer and 24.5 s median latency make the cost-quality trade-off explicit: deep research costs 3.4 times single-shot RAG for 14 points of BioASQ correctness and the highest answer relevancy, whereas Corrective RAG gives the cleanest evidence at 1.8 times the cost. A deployment can choose the strategy by question type, for example the food-data path (6,356 tokens, 10.4 s) for nutrient questions and deep research for contested evidence.

Text exact match was 0 and BLEU about 0.06 even for the method whose BioASQ answers were judged correct nearly three times out of four, and BERTScore varied by only 0.02 across methods. For explanatory health answers, reference-overlap metrics should be reported with, not instead of, judged correctness and claim-level grounding. Conversely, the recipe-chatbot study found that RAGAS faithfulness and relevancy did not predict user ratings~\cite{jeong2026}, so automatic grounding metrics should in turn be complemented by user evaluation.

\section{Limitations and future work}\label{sec:5}

The answering model, relevance judge and BioASQ grader are Claude Sonnet, and the claim judge is Claude Opus; LLM judgments correlate with human assessment but do not replace it~\cite{clarke2024,faggioli2023,zheng2023}, and models of one family may share blind spots. The conversational evaluation is small (36 turns in 12 scripted conversations, 30 memory-depth conversations, 38 load-test questions) and written by the researchers, so it shows behavior and validates the metrics rather than estimating performance on real users' questions; the topic-leakage estimate rests on three switches. Faithfulness and answer relevancy for the RAG methods were judged on 200 of the 1,000 questions per method. Two metric caveats apply: a claim labeled ``unsupported'' was not found in the provided sources but may still be true, and overlap metrics depend on reference wording, with NFCorpus and SciFact providing no free-text references. Costs are list-price estimates, whereas the study used a subscription, and the load test ran on one workstation. Gold mapping relies on DOIs and titles, and many OpenAlex records lack abstracts; OpenAlex assigns each work one machine-predicted primary topic, so the slice sizes in Supplementary Table~\ref{tab:S1} depend on that classifier and on our topic list.  Only one answering model was evaluated. SHAKTI gives research information, not medical advice.

In future studies, we will (i) run a human study of answer usefulness, trust and readability with the deployed system, after ethics approval, with SHAKTI's three evidence strategies assigned at random and blinded; (ii) validate the claim judge against expert annotation of a claim sample and report agreement; (iii) evaluate the methods on the reranked retrieval and on all 10,065 questions, with lexical retrieval and a citation-graph RAG over OpenAlex; (iv) replicate with answering models and judges from other model families; (v) add personalization (dietary restrictions, conditions, cultural cuisines)   questions; (vi) recover missing abstracts from PubMed; and (vii) reduce latency with caching and streaming, and route questions to the cheapest adequate evidence strategy.

\section{Conclusion}\label{sec:6}

SHAKTI shows that a nutrition and health assistant can answer from the whole scholarly record and from exact food data while keeping its claims auditable. Hybrid retrieval with MedCPT reranking found benchmark evidence best; agentic RAG selected cleaner and more relevant evidence and answered 73\% of BioASQ questions correctly; across all configurations faithfulness was about 95-96\%, fewer than 2\% of claims were unsupported and almost none contradicted the sources. In conversation SHAKTI kept the context of follow-ups (context-awareness score 97.1\%), remembered the first question after 16 unrelated ones, cited only working sources and quoted them verbatim, at about six US cents and a median of 24 seconds per answer. Closed-form verdict accuracy remained moderate, and reference-overlap metrics did not reflect judged correctness. Health RAG systems should therefore be evaluated across retrieval, generation, grounding, efficiency and conversation together, with mechanical verification of every cited claim, and validated with users.

\section*{CRediT authorship contribution statement}

Chaitanya Pallerla: Conceptualization, Methodology, Software, Investigation, Data curation, Formal analysis, Writing - original draft.

Dongyi Wang: Writing - review and editing.

\section*{Declaration of competing interest}

The authors declare that they have no known competing financial interests or personal relationships that could have appeared to influence the work reported in this paper.

\section*{Funding}

This research did not receive any specific grant from funding agencies in the public, commercial, or not-for-profit sectors.

\section*{Ethics statement}

This study used publicly available benchmark datasets and bibliographic metadata. All conversational tests were scripted and run by the researchers; no human participants or personal data were involved. The planned user study will be submitted for ethics review before recruitment.

\section*{Data availability}

The benchmarks (NFCorpus, BioASQ, SciFact, PubMedQA), the OpenAlex snapshot and USDA FoodData Central are publicly available. The SHAKTI source code, including the indexing, retrieval, RAG and evaluation scripts, is available at \href{https://github.com/Chaitanya-Pallerla/shakti-rag-nutrition}{\textit{Github Link}} and archived at \href{https://doi.org/10.5281/zenodo.24423092}{\textit{Zenodo link}}. The question set with gold-standard mappings, rankings, LLM outputs, relevance and claim-level judgments, and the scripted conversations are available from the corresponding author upon reasonable request.

\section*{Declaration of generative AI and AI-assisted technologies in the writing process}

During the preparation of this work, the authors used Claude (Anthropic) to improve grammatical accuracy, sentence structure, and overall clarity of the manuscript. The authors subsequently reviewed and edited the content as necessary and take full responsibility for the content of the published article.  Separately, Claude Sonnet and Claude Opus are components of the system and of the evaluation (answering model, relevance judge, answer grader and claim judge), as described in Section~\ref{sec:2}.

\bibliographystyle{IEEEtran}
\bibliography{refs}

\clearpage
\setcounter{table}{0}
\renewcommand{\thetable}{S\arabic{table}}
\renewcommand{\theHtable}{S\arabic{table}}

\begin{table*}[!t]
\centering
{\scshape Supplementary Material\par}\medskip
\caption{Nutrition-related literature in the corpus and where the relevant evidence sits.}
\label{tab:S1}
\par\smallskip{\footnotesize\textit{Part A. Corpus slices (counts exclude paratext and retracted works; about 490 million records).}}
\par\medskip
{\footnotesize\setlength{\tabcolsep}{4pt}
\begin{tabular}{@{}L{0.495\linewidth}C{0.306\linewidth}C{0.153\linewidth}@{}}
\toprule
\textbf{Slice} & \textbf{Works} & \textbf{Reviews} \\
\midrule
Nutrition and Dietetics + Food Science subfields & 3.10 M (1.08 M + 2.01 M) & 61,227 \\
Above + 104 nutrition/diet/food topics in other fields & 7.56 M & 191,640 \\
Health, life-science and agriculture fields* & 114.3 M & n/a \\
No topic assigned & 102.9 M & n/a \\
\bottomrule
\end{tabular}}
\par\medskip
{\scriptsize *Medicine, Nursing, Health Professions, Dentistry, Pharmacology/Toxicology/Pharmaceutics, Immunology and Microbiology, Biochemistry/Genetics/Molecular Biology, Neuroscience, Veterinary, Agricultural and Biological Sciences.}
\par\medskip
\par\smallskip{\footnotesize\textit{Part B. Share of gold (relevant) papers that each slice still contains.}}
\par\medskip
{\footnotesize\setlength{\tabcolsep}{4pt}
\begin{tabular}{@{}L{0.148\linewidth}C{0.128\linewidth}C{0.192\linewidth}C{0.213\linewidth}C{0.241\linewidth}@{}}
\toprule
\textbf{Benchmark} & \textbf{Gold items} & \textbf{Nutrition slice (7.56 M)} & \textbf{Health/life/agri slice (114.3 M)} & \textbf{Questions with $\geq$1 gold left, nutrition slice} \\
\midrule
NFCorpus & 114,638 & 28.9\% & 87.5\% & 82.8\% \\
BioASQ & 42,267 & 2.5\% & 96.7\% & 7.6\% \\
PubMedQA & 999 & 2.6\% & 91.7\% & 2.6\% \\
SciFact & 1,134 & 6.2\% & 95.6\% & 5.8\% \\
All & 159,038 & 21.6\% & 90.0\% & 31.0\% \\
\bottomrule
\end{tabular}}
\par\medskip
{\scriptsize Gold items are counted per question; every gold paper has an OpenAlex topic (0\% are untagged). Of the 3,084 distinct NFCorpus gold papers, 5.9\% are in Nutrition and Dietetics and 2.0\% in Food Science; by field, 54.5\% are in Medicine, 12.3\% Biochemistry/Genetics/Molecular Biology, 7.6\% Environmental Science (mainly Health, Toxicology and Mutagenesis), 6.9\% Agricultural and Biological Sciences and 5.9\% Nursing. The 104-topic list is provided as a supplementary file (nutrition\_topics.csv). The list is keyword-based and includes a few borderline topics (e.g. soil/plant nutrients, calorimetry), so 7.56 M is an upper estimate.}
\end{table*}

\end{document}